\documentclass{article}

\usepackage[preprint]{neurips_2023}
\setcitestyle{numbers,square}

\usepackage[utf8]{inputenc} 
\usepackage[T1]{fontenc}    
\usepackage[colorlinks,
            linkcolor=pink,       
            anchorcolor=blue,  
            citecolor=blue,        
            ]{hyperref}     
\usepackage{url}            
\usepackage{booktabs}       
\usepackage{amsfonts}       
\usepackage{nicefrac}       
\usepackage{microtype}      
\usepackage{xcolor}         
\usepackage{natbib}
\usepackage{tabularx}

\usepackage{amsmath,amssymb,amsfonts}
\usepackage{algorithm}
\usepackage{algorithmic}
\usepackage{graphicx}
\usepackage{textcomp}
\usepackage{multirow} 
\usepackage{makecell}
\usepackage{bbm}
\usepackage{mathtools}
\usepackage{bbold}
\usepackage{caption}

\newcommand{\ie}{\textit{i.e.}, }
\newcommand{\eg}{\textit{e.g.}, }

\title{X-IQE: eXplainable Image Quality Evaluation \\
for Text-to-Image Generation \\
with Visual Large Language Models}

\author{%
  Yixiong Chen\\
  CUHK-SZ, SRIBD\\
  \texttt{yixiongchen@link.cuhk.edu.cn} \\
  \And
  Li Liu\thanks{Corresponding Author} \\
  HKUST-GZ \\
  \texttt{avrillliu@hkust-gz.edu.cn} \\
  \AND
  Chris Ding \\
  CUHK-SZ \\
  \texttt{chrisding@cuhk.edu.cn} \\
}

\begin{document}

\maketitle

\begin{abstract}
This paper introduces a novel explainable image quality evaluation approach called X-IQE, which leverages visual large language models (LLMs) to evaluate text-to-image generation methods by generating textual explanations. X-IQE utilizes a hierarchical Chain of Thought (CoT) to enable MiniGPT-4 to produce self-consistent, unbiased texts that are highly correlated with human evaluation. It offers several advantages, including the ability to distinguish between real and generated images, evaluate text-image alignment, and assess image aesthetics without requiring model training or fine-tuning. X-IQE is more cost-effective and efficient compared to human evaluation, while significantly enhancing the transparency and explainability of deep image quality evaluation models.
We validate the effectiveness of our method as a benchmark using images generated by prevalent diffusion models. X-IQE demonstrates similar performance to state-of-the-art (SOTA) evaluation methods on COCO Caption, while overcoming the limitations of previous evaluation models on DrawBench, particularly in handling ambiguous generation prompts and text recognition in generated images. Project website: \href{https://github.com/Schuture/Benchmarking-Awesome-Diffusion-Models}{https://github.com/Schuture/Benchmarking-Awesome-Diffusion-Models}
\end{abstract}

\section{Introduction}
Image quality evaluation has long been a practical and crucial technique employed in various applications, including photo enhancement~\cite{bhattacharya2011holistic}, image stream ranking~\cite{chang2017aesthetic}, and album thumbnail composition~\cite{chen2017learning}. With the advent of artificial intelligence (AI) and generative models, such as diffusion models~\cite{song2020score,nichol2021improved}, there is an increasing demand for effective evaluation methods to assess the large volume of images generated. The evaluation process of image quality by the human visual system encompasses multiple factors, such as the rationality of the image content, the alignment with text descriptions~\cite{wu2023better}, and aesthetics~\cite{sheng2018attention}. However, some of these evaluation factors are inherently subjective and challenging to quantify, presenting a significant obstacle in this task.

Existing solutions for image quality evaluation can be broadly categorized into two groups: human evaluation and model evaluation. Human evaluation, despite its widespread use, suffers from significant drawbacks, such as high cost (\eg crowd-sourcing) and limited reproducibility across different evaluation groups~\cite{otani2023toward}. On the other hand, model evaluation provides a more desirable alternative; however, it often relies on complex models, including CNN-LSTM~\cite{chang2017aesthetic}, CLIP~\cite{radford2021learning}, and BLIP~\cite{li2022blip}, along with specially labeled data and features. This is also expensive and falls short of the strong generalization capabilities of humans (\eg on AI-generated images). Furthermore, most previous model-based evaluations only focus on predicting image quality scores~\cite{sheng2018attention,xu2023imagereward}, which makes it challenging to explain the biases and deficiencies in their training data, ultimately resulting in flawed model performance. Consequently, the question of how to develop cheap, generalizable, and explainable quality evaluation models remains an open challenge.

\begin{figure}[t]
\vspace{0cm}                          
\centering\centerline{\includegraphics[width=1.0\linewidth]{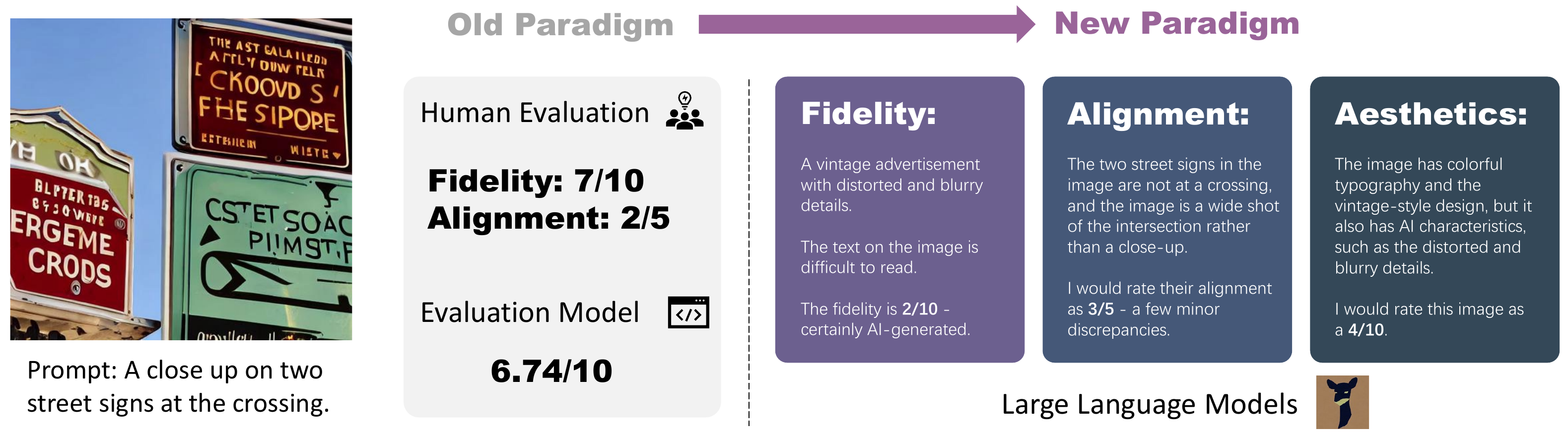}}
\caption{The paradigm shift from human/model evaluation to LLM explanation of image quality.}
\label{fig:motivation}  
\end{figure}

In this work, we propose a novel approach that leverages pre-trained visual large language models (LLMs) to generate analysis texts for images in a conversational style (see Fig. \ref{fig:motivation}). We utilize the state-of-the-art (SOTA) visual LLM, MiniGPT-4~\cite{zhu2023minigpt}, as our foundational model for implementing explainable image quality evaluation. Leveraging the in-context learning (ICL) capability~\cite{brown2020language,min2022rethinking} of LLMs, we can inject knowledge into them without further training. Specifically, we propose an effective prompt set as the standard for LLMs to evaluate image quality. Our method addresses several key drawbacks associated with existing image quality evaluation models. The advantages of our proposed explainable image quality evaluation method, X-IQE, are as follows:

\begin{itemize}
    \item \textbf{Explainability}: Models can describe their reasoning process using Chain of Thought (CoT)~\cite{weichain} based on the provided prompts, making the results explainable and insightful.
    \item \textbf{Comprehensiveness}: Carefully designed prompts allow LLMs to perform comprehensive evaluations, not only for specifically labeled features (\eg color, composition~\cite{lee2023aligning,zhang2021image}).
    \item \textbf{Powerful Performance}: Advanced LLMs are trained on vast amounts of data and possess greater generalizable image understanding capabilities than most task-specific models.
    \item \textbf{Unbiasedness}: By utilizing objective prompt text, our model can conduct unbiased evaluations, eliminating the biases that may arise from models trained on datasets annotated by specific annotation groups, such as crowd-sourcing~\cite{otani2023toward} or annotation companies~\cite{xu2023imagereward}.
    \item \textbf{Training-free}: Our method harnesses the capabilities of pre-trained LLMs, eliminating the need for data collection and training efforts required by existing methods~\cite{lee2023aligning,xu2023imagereward}.
\end{itemize}

Recent studies for evaluating AI-generated images~\cite{xu2023imagereward,wu2023better,otani2023toward}have predominantly concentrated on assessing fidelity and text-image alignment scores. Conversely, earlier works assessing real photos or paintings have primarily emphasized aesthetics~\cite{chang2017aesthetic,sheng2018attention}. In contrast to existing methods that directly output evaluation scores, we establish a step-by-step text-to-image method evaluation pipeline. This approach enables us to generate all three textual evaluations incrementally, allowing subsequent steps to leverage the information and conclusions derived from the reasoning of previous steps.

To provide a comprehensive analysis of X-IQE and evaluate SOTA text-to-image generative models, including Stable Diffusion~\cite{rombach2022high}, Openjourney~\footnote{https://openjourney.art/}, and DeepFloyd-IF~\footnote{https://huggingface.co/DeepFloyd/IF-I-XL-v1.0}. we conduct extensive experiments. Remarkably, X-IQE performs comparably in a zero-shot manner with specialized SOTA scoring models trained on the AI-generated images or even better in unusual scenarios. These results demonstrate the efficacy of our method as a versatile text-to-image evaluation protocol.

The contributions of this work are summarized as follows:

\begin{enumerate}
    \item We propose X-IQE, an explainable image quality evaluation method based on visual LLMs. To our knowledge, this is the first application of LLMs for text-to-image evaluation.
    \item To enhance the performance and stability of X-IQE, we integrate the expertise of art professionals into a hierarchical CoT with well-defined conditions and criteria for evaluation.
    \item We perform comprehensive experiments on both real and AI-generated images, validating the explanatory power as well as the quantitative scoring capabilities of X-IQE.
\end{enumerate}

\section{Related Work}
\label{related_work}

\subsection{Image Quality Evaluation for Text-to-Image Generation}

Human evaluation is widely regarded as the benchmarking method for assessing text-to-image generative models, including rule-based methods~\cite{hu2023label}, GANs~\cite{goodfellow2020generative}, and diffusion models~\cite{ho2020denoising}, adopted by most SOTA text-to-image generative models~\cite{ding2022cogview2,yu2022scaling,nichol2022glide,saharia2022photorealistic}. However, the lack of a consistent evaluation standard and varying protocols often yield different conclusions among the works.
To overcome this limitation, the first approach involves employing standardized evaluation protocols, which addresses challenges such as prioritizing monetary returns~\cite{aguinis2021mturk,kennedy2020shape} and introducing biases~\cite{otani2023toward}. The second approach focuses on automatic evaluation metrics, including Inception Score~\cite{salimans2016improved}, Frechet Inception Distance~\cite{heusel2017gans}, and Precision-Recall~\cite{kynkaanniemi2019improved}. Some recent studies have trained evaluation models using human scoring to align their preferences, but achieving only 65\% consistency~\cite{xu2023imagereward,wu2023better}. Notably, both approaches consider two metrics: overall image quality and text-image alignment.

\subsection{Image Aesthetic Analysis}
Prior aesthetic prediction models have predominantly focused on real images, employing various approaches for aesthetic assessment. One common practice involves concatenating vector representations derived from multiple input image patches~\cite{lu2015deep,sheng2018attention}. Another approach aims to capture the relationships between different objects or regions within the image to evaluate composition~\cite{liu2020composition,zhang2021image}. Notably, the widely used text-to-image generation method, Stable Diffusion, has also been evaluated using aesthetic predictors trained on AVA~\cite{murray2012ava} and LAION~\cite{schuhmannlaion} datasets, yielding favorable results.

\subsection{Large Language Models}
Large language models (LLMs) have achieved remarkable success in recent years. Three main paradigms emerged: encoder-only (BERT~\cite{devlin2018bert}), encoder-decoder (T5~\cite{radford2019language}), and decoder-only (GPT-2~\cite{raffel2020exploring}). GPT-3~\cite{brown2020language} demonstrated the scaling advantages of decoder-only paradigm, leading to a surge in related research, including Megatron-Turing NLG~\cite{smith2022using}, Chinchilla~\cite{hoffmann2022training}, PaLM~\cite{chowdhery2022palm}, and LLaMA~\cite{touvron2023llama}. InstructGPT~\cite{ouyang2022training} and ChatGPT~\cite{bubeck2023sparks} showed coherent multi-turn conversation skills through fine-tuning GPT-3 with aligned feedback data.
LLMs can generate expected outputs for test instances without additional training when provided with natural language instructions and/or task demonstrations~\cite{zhao2023survey}. This in-context learning ability~\cite{brown2020language} allows LLMs to learn new tasks with minimal overhead during inference. Another important capability is Chain of Thought, where LLMs solve tasks using a prompting mechanism involving intermediate reasoning steps. CoT prompting has shown performance gains for models larger than 60B~\cite{weichain}. In this work, we leverage ICL and CoT to enable smaller LLMs to explain their reasoning process in image quality evaluation.

\subsection{Pre-trained LLMs in Visual-Language Tasks}
LLMs, such as VisualGPT~\cite{chen2022visualgpt} and Frozen~\cite{tsimpoukelli2021multimodal}, serve as powerful decoders for visual features. They enable cross-modal transfer, aligning visual and linguistic knowledge to describe visual information using language. BLIP-2~\cite{li2023blip} effectively aligns visual features with language models through Flan-T5~\cite{chung2022scaling}, demonstrating strong visual QA capabilities. GPT-4~\cite{bubeck2023sparks}, a recent breakthrough, accomplishes diverse language tasks based on images by aligning an advanced LLM with human preferences and intentions. Successful visual-language conversation models require robust conversational language models (\eg ChatGPT, LLaMA~\cite{touvron2023llama}, Vicuna~\cite{chiang2023vicuna}), visual encoders (\eg VIT~\cite{dosovitskiy2020image}), and visual-language alignment training. MiniGPT-4~\cite{zhu2023minigpt}, chosen for its versatility and simplicity, achieves dialogue-based image understanding by training a linear layer with ample image-text alignment data. Our proposed method can be flexibly adapted to more advanced visual language models in the future.

\section{Method}
\label{method}

\begin{figure}[!htbp]                    
\centering\centerline{\includegraphics[width=1.0\linewidth]{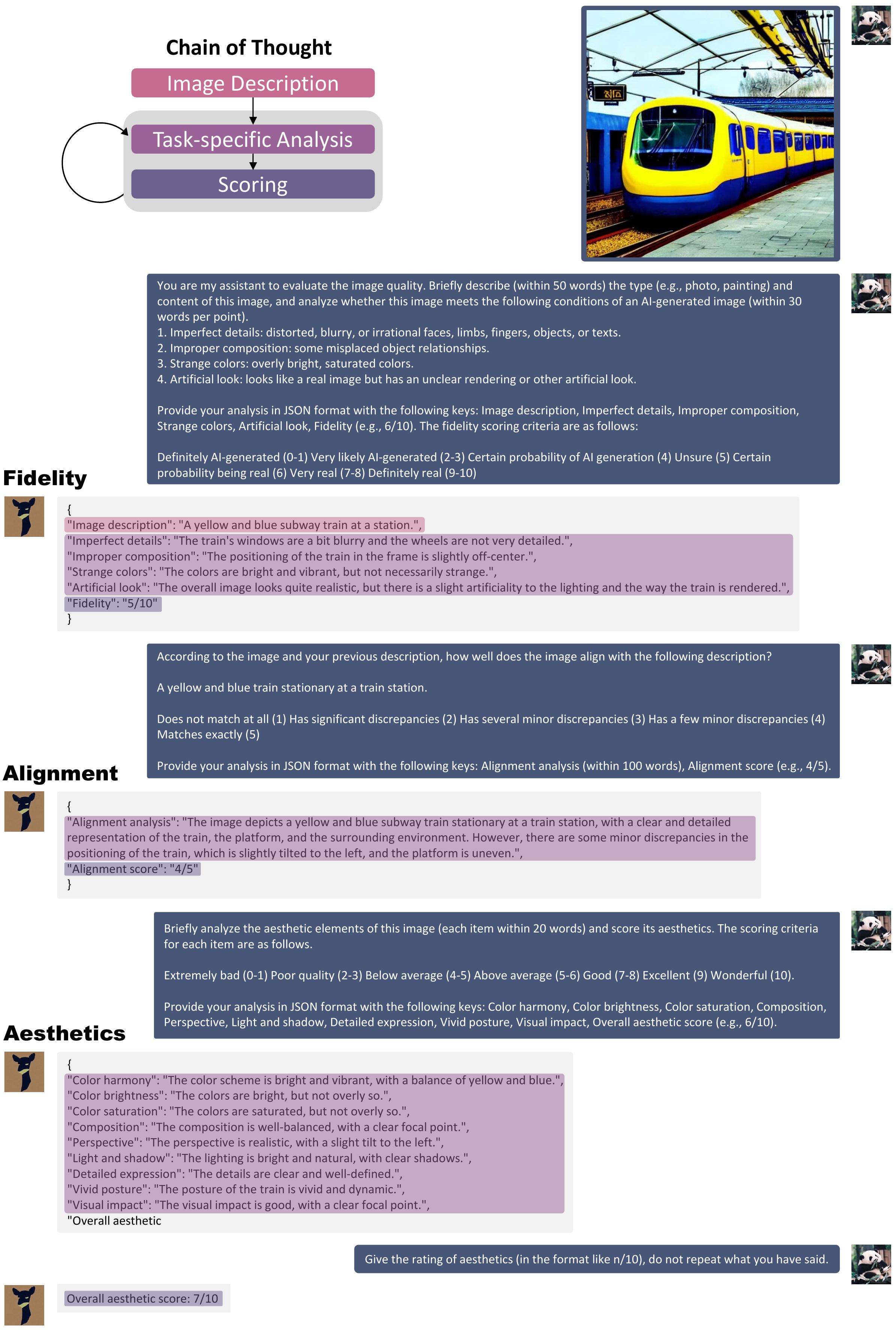}}
\caption{The illustration showcases X-IQE using an image generated by Stable Diffusion 1.4. In X-IQE, MiniGPT-4 with Vicuna serves as the chatbot. Carefully designed prompts are employed for image quality analysis, incorporating a hierarchical CoT strategy to ensure response logic and accuracy. The output format and scoring criteria constraints contribute to stable and coherent responses that are easy to follow.}
\label{fig:method}
\vspace{-0.5cm}
\end{figure}
X-IQE utilizes MiniGPT-4 as its foundational model and incorporates the expertise of art professionals to design standardized and unbiased prompts. The effectiveness of X-IQE is further enhanced by its well-structured hierarchical CoT scheme and the application of format constraints for accurate execution. The overall methodology is depicted in Fig. \ref{fig:method}.

\subsection{MiniGPT-4}
MiniGPT-4~\cite{zhu2023minigpt} combines a pretrained vision encoder (ViT with Q-Former~\cite{li2023blip}) and the advanced LLM Vicuna~\cite{chiang2023vicuna}. ViT enables image parsing, while Vicuna enhances generative capabilities through conversation rounds. MiniGPT-4 uses frozen ViT and Vicuna, with training involving a linear projection from visual features to Vicuna. The model is trained on 5 million text-image pairs for general vision-language knowledge and fine-tuned with 3.5k aligned conversations for natural responses. In this work, we perform ICL on the frozen MiniGPT-4 for image quality evaluation.

\textbf{Challenges.} Using pretrained LLMs for image quality assessment presents challenges that require careful ICL strategy design. First, the training of MiniGPT-4 lacks samples from SOTA visual generative models. So LLMs tend to describe them as normal images based on pre-training preferences. We address this challenge by incorporating explicit quality evaluation aspects, standards, and reasoning processes inspired by art professionals' knowledge. Additionally, Vicuna in MiniGPT-4 has a limited parameter count (7B/13B), which may affect analysis rationality and result accuracy in complex scenarios. To mitigate this, we impose strict constraints on the answer structure.

\subsection{Expertise from Art Professionals for ICL}
Our art industry consultant, experienced in using Stable Diffusion and Midjourney, identified several empirical discrimination methods for AI image generation. When generating rare and complex objects, AI produces blurred or distorted details, particularly with human hands and text. Occluded objects often have inconsistent details on both sides of the occluder. AI-generated images may exhibit oversaturation, especially when unusual colors are present. Lastly, in photo scenes, AI-generated images can sometimes have a partially greasy feeling.

Aesthetics judgment indicators are well-studied and less subjective than commonly believed. Aesthetically pleasing images should exhibit harmonious and bright colors, moderate saturation, appropriate lighting and shadow, well-composed scenes, and a sense of space through perspective~\cite{palmer2013visual}. Memorable images should also feature intricate details, expressiveness, and strong visual impact~\cite{khosla2015understanding}. X-IQE incorporates expertise for discriminating AI images and aesthetic evaluation as prompts.

\subsection{Chains of Thought between and within Tasks}

We design a hierarchical CoT strategy for X-IQE to enhance the coherence and quality of its responses. X-IQE evaluates image quality in terms of fidelity, alignment, and aesthetics, recognizing that these attributes are interconnected. The evaluation results of certain attributes can serve as priors for assessing other attributes. Fidelity, which measures the realism of an image, influences the assessment of text-to-image alignment since AI-generated images often exhibit lower alignment compared to real images~\cite{otani2023toward}. Additionally, the identification of an image as AI-generated impacts the aesthetic analysis by highlighting specific defects identified in the fidelity evaluation. The alignment evaluation benefits from knowing the true prompt used for image generation, aiding in the assessment of visual impact and composition in the aesthetic evaluation. Our CoT is designed as follows:

\begin{itemize}
    \item $fidelity~evaluation~~\longrightarrow~~alignment~evaluation~~\longrightarrow~~aesthetic~evaluation$
\end{itemize}

Within each evaluation task, a dedicated CoT is established, encompassing:

\begin{itemize}
    \item $(image~description)~~\longrightarrow~~task~specific~analysis~~\longrightarrow~~scoring$
\end{itemize}

The image description is included in the first task and reused in the subsequent tasks.

\subsection{Constraints of Output Formats and Scoring Criteria}

During our preliminary experiments, we aimed to ensure that MiniGPT-4 produces content that adheres to the CoT structure. However, we encountered issues with unstable outputs. The two most common unexpected outputs were: 1) content being generated before the analysis and 2) scoring presented in varying styles, such as percentages or ratings based on different scales. To enhance the likelihood of CoT-compliant output and establish an objective and unified scoring standard, we implemented two improvements: 1) requiring the model to provide output in JSON format and 2) explicitly defining the conditions associated with different scores.

\section{Experiments and Results}

\subsection{Experimental Settings}
\textbf{Datasets:} To evaluate X-IQE's capability, we utilize COCO Captions~\cite{chen2015microsoft} and DrawBench texts~\cite{saharia2022photorealistic} for AI image generation. We randomly sample 1000 prompts from COCO Captions while all 200 prompts from DrawBench are used. All diffusion models generate images using DPM Solver~\cite{ludpm} with 25 steps and a guidance scale of 7.5.

\textbf{Models:} Our experiments involve both MiniGPT-4 models with 7B and 13B Vicuna parameters. No major modifications are made to the model code base\footnote{https://github.com/Vision-CAIR/MiniGPT-4}. During inference, we set beam=1, and the temperature ranges from 0.01 to 1.0. The models are executed on an Nvidia RTX A6000 48G GPU.

\textbf{Metrics:} To validate X-IQE's performance in fidelity evaluation, we employ the recall of the AI-generated images. For alignment and aesthetics validation, we compare the Pearson correlation coefficient between human evaluation and task-specific models (such as CLIPScore~\cite{hessel2021clipscore} for alignment and Stable Diffusion Aesthetic Predictor\footnote{https://github.com/christophschuhmann/improved-aesthetic-predictor} for aesthetics) or X-IQE.

\subsection{The Impact of Model Size and Temperature}

\begin{figure}[!htbp]                    
\centering\centerline{\includegraphics[width=1.0\linewidth]{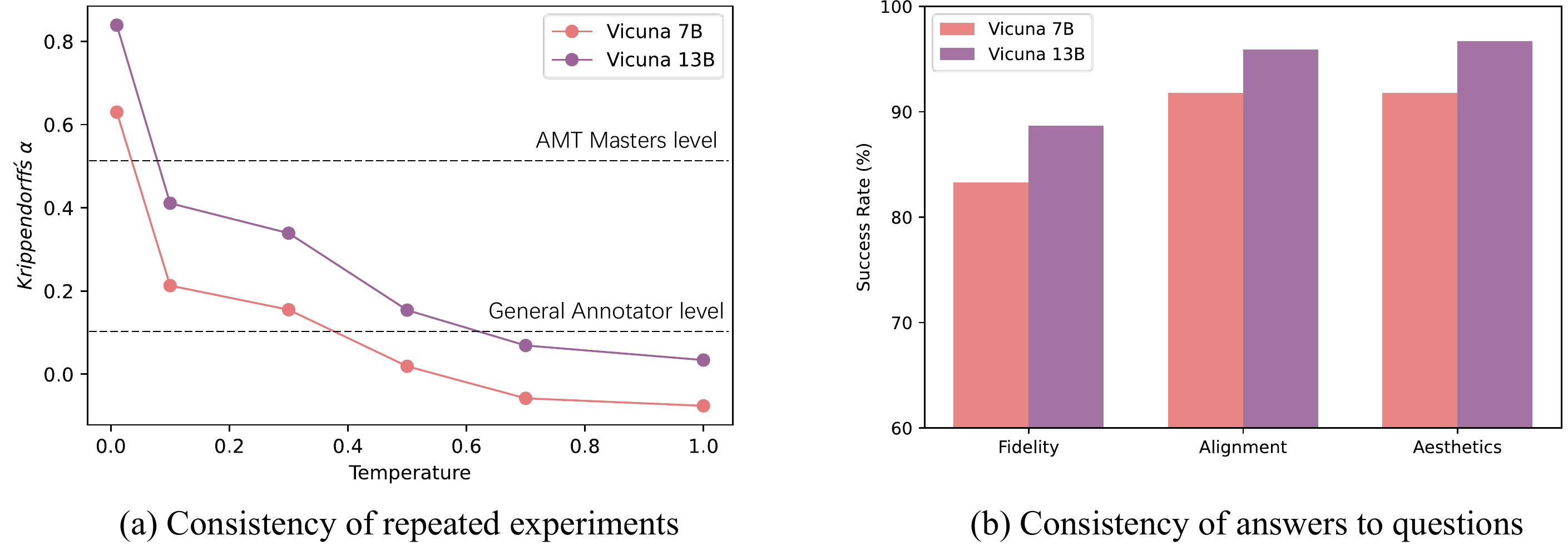}}
\caption{The stability analysis of X-IQE. (a) the models output more consistent results when the temperature is lower, with Vicuna 13B always better than the 7B variant. (b) Larger LLM gives more precise and valid response to the questions for all three tasks.}
\label{fig:consistency}
\end{figure}

One question regarding LLMs is whether multiple experiments yield consistent judgments like the traditional prediction models. And if its responses are always consistent with our questions. We investigate model consistency in relation to model size and temperature, as depicted in Fig. \ref{fig:consistency}.

The temperature parameter is crucial in LLMs as it controls the randomness and creativity of generated text. We conduct three repeated experiments with fixed temperatures to examine the consistency of models' fidelity prediction (Fig. \ref{fig:consistency} (a)). Krippendorff's $\alpha$~\cite{krippendorff2018content} decreases as the temperature increases, indicating that models yield more divergent results with higher temperatures. General annotator groups typically achieve 0.11 Krippendorff's $\alpha$, while expert annotators can reach 0.53~\cite{otani2023toward}. Moreover, Vicuna 13B consistently exhibits better consistency than Vicuna 7B.

X-IQE aims to derive quantitative evaluations for three tasks, but the models do not always provide numerical results as expected. For fidelity evaluation, a common response is \textit{"It's a bit unclear if the image is AI-generated or not"}, even after two rounds of questioning \textit{"Give the rating of fidelity (in the format like n/10)"}. The success rates of answering are presented in Fig. \ref{fig:consistency} (b). Generally, larger Vicuna performs better in this experiment. Both models have a probability over 10\% of providing unexpected fidelity answers. However, refusal to answer for alignment and aesthetics evaluation is rare, with Vicuna 13B inconsistently answering questions only in exceptional cases.

In all subsequent experiments, we employ Vicuna 13B with a temperature of 0.1 for accuracy and reproducibility. Smaller temperatures are avoided due to model instability, as discussed in Section \ref{sect:limitations}.

\subsection{The Role of CoT}

\begin{table}[t]
\small
\centering
\caption{Ablation study of CoTs within and between tasks. The baseline is directly asking MiniGPT-4 for separate evaluation scores. Fidelity is measured using the recall of generated images, while alignment and aesthetics are assessed using Pearson correlation coefficients with human evaluations. It's worth noting that the human evaluation exhibits a correlation coefficient of 0.137 with the CLIPScore and 0.067 with the Aesthetic Predictor, serving as a reference for comparison.}
\setlength{\tabcolsep}{0.4mm} 
\renewcommand{\arraystretch}{1.3} 
\begin{tabular}{c|c|c|c|c|c|c}
\toprule
\multicolumn{1}{c|}{\multirow{2}{*}{Subtask}} &  \multicolumn{3}{c|}{CoT within tasks} & \multicolumn{3}{c}{CoT between tasks} \\
\multicolumn{1}{c|}{} & ~~baseline~~ & ~~~+prompt~~~ & ~~+prompt+format~~ & ~+fidelity~ & ~~+alignment~~ & ~+fidelity+alignment~  \\
\hline
\hline
\rule{0pt}{4pt} 
~~Fidelity~~ & 0.021 & 0.0 & 0.698 & -- & -- & -- \\
~~Alignment~~ & 0.118 & 0.082 & 0.263 & 0.381 & -- & -- \\
~~Aesthetics~~ & 0.030 & -0.162 & 0.259 & 0.369 & 0.351 & 0.418 \\
\bottomrule
\end{tabular}
\label{tab:cot}
\end{table}

The ablation study results of CoT are shown in Tab. \ref{tab:cot}. We observe that without CoT within tasks, the performance of X-IQE is unsatisfactory. Simply asking the model for a fidelity or alignment score typically yields high scores, as no consistent standards are provided. Incorporating specific requirements for task-specific analysis, including scoring criteria and evaluation aspects, improves the model's ability to generate reasonable analysis. However, obtaining reliable numerical scores remains challenging due to Vicuna's tendency to provide answers before thorough analysis. The inclusion of formatting requirements ultimately enables the model to produce desirable quantitative results based on its analysis. For the results of CoT between tasks, the alignment and aesthetics scoring can match the human evaluations better with the analysis of previous stages. The model frequently utilizes earlier analysis when responding to subsequent questions. For example, fidelity analysis considering image details often aids in evaluating aesthetics such as composition and expressive details.

\subsection{Real or Fake? Fidelity Evaluation}
\label{sect:fidelity}

\begin{figure}[!tbp]
\vspace{0cm}                          
\centering\centerline{\includegraphics[width=1.0\linewidth]{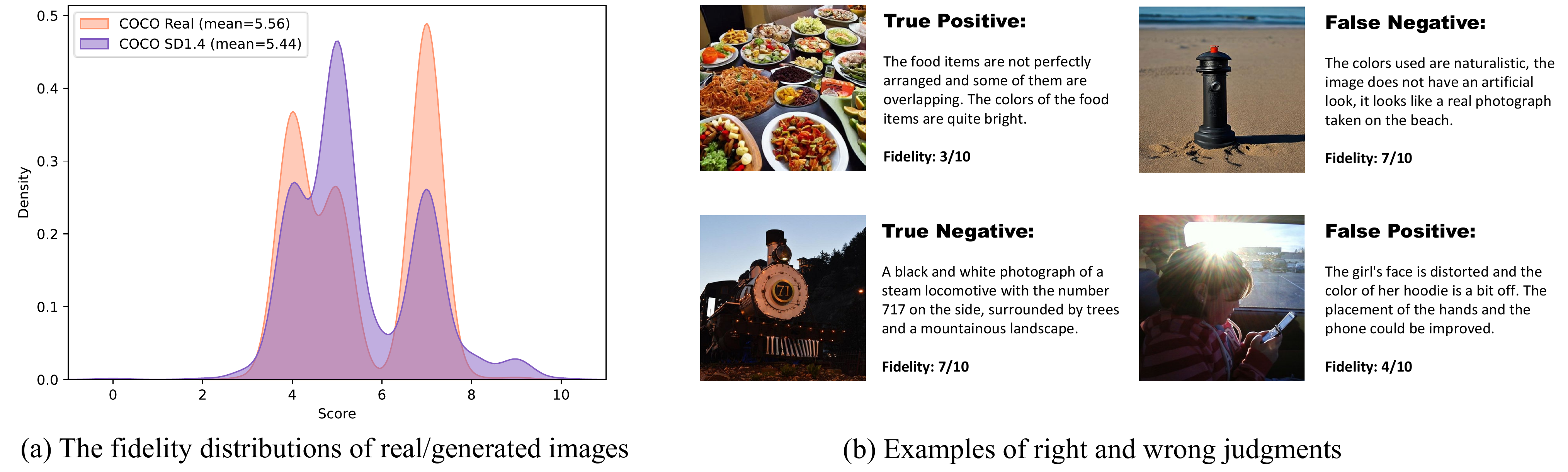}}
\caption{The results and examples of fidelity evaluation on COCO Caption. (a) Real and fake images mainly differ in the distribution of fidelity scores. (b) The examples illustrate the capability of X-IQE to perform image analysis during inference, although some judgments may be inaccurate.}
\label{fig:fidelity}  
\end{figure}

The results and examples of fidelity inference using X-IQE are presented in Fig. \ref{fig:fidelity}. Though the mean fidelity scores for real and generated images are similar, it does not imply that X-IQE lacks strong discriminative ability. The fidelity score distributions of real and generated images (Fig. \ref{fig:fidelity} (a)) exhibit significant differences with a p-value of $<10^{-5}$ in the Kolmogorov–Smirnov test. X-IQE tends to assign more scores of 7 to real images and more scores of 5 to SD-generated images. Interestingly, SD 1.4 can deceive both X-IQE and human observers in quite a few cases. Fig. \ref{fig:fidelity} (b) demonstrates examples of correct and incorrect judgments for detecting AI generation. Furthermore, X-IQE assigns very few scores of 6, possibly due to the language bias learned during pretraining.

\subsection{Alignment and Aesthetics Evaluations}
\label{sect:align_aes}

\begin{figure}[t]                         
\centering\centerline{\includegraphics[width=0.8\linewidth]{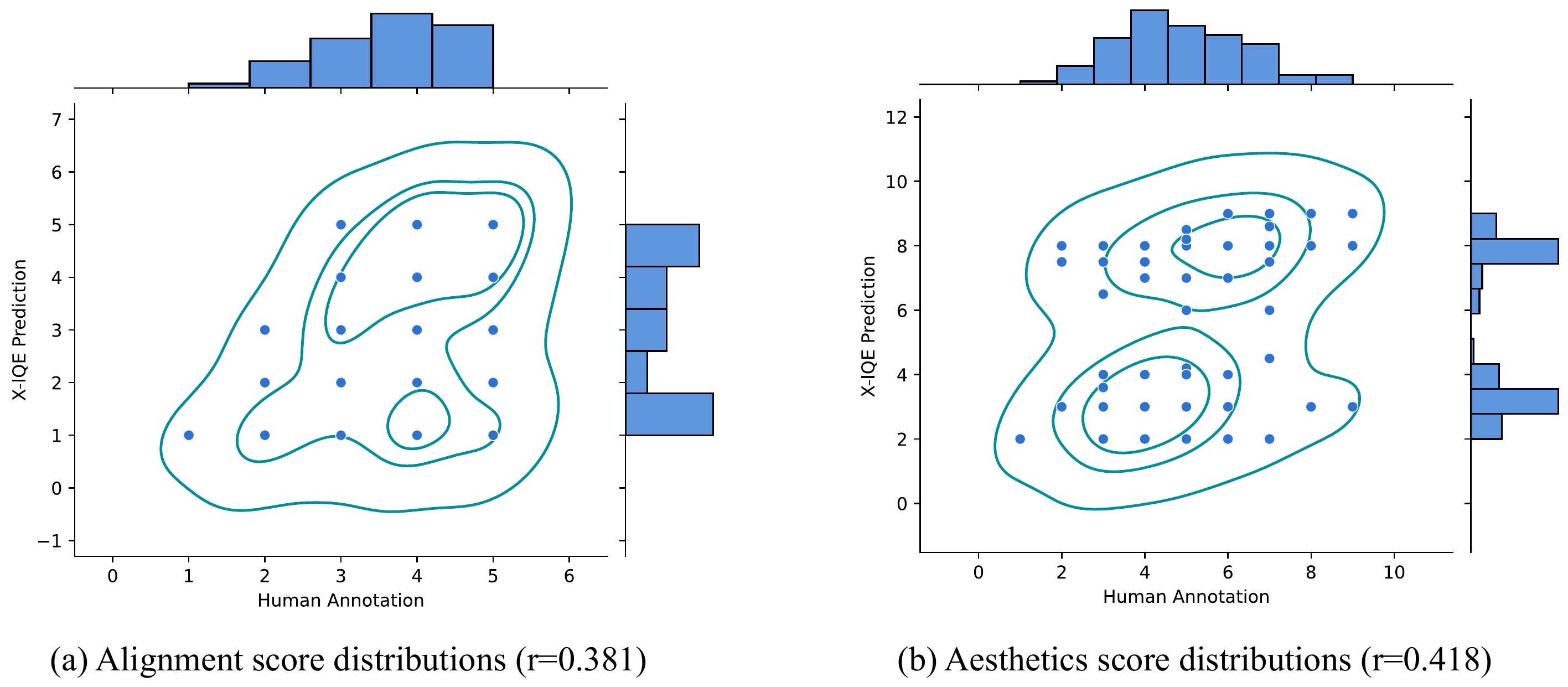}}
\caption{The distributions of alignment and aesthetics scores on COCO Caption. In contrast to humans, who tend to provide score judgments centered around the mean, X-IQE exhibits a broader range of judgments that include both extremely high and extremely low values.}
\label{fig:align_aes} 
\end{figure}

The distributions of alignment and aesthetics scores from X-IQE and human evaluations are depicted in Fig. \ref{fig:align_aes}. These scores demonstrate higher correlation coefficients of 0.381 and 0.418 with human evaluations compared to task-specific models like CLIPScore (0.137) and Aesthetic Predictor (0.067), highlighting the generalizability of X-IQE on the generated dataset. CLIPScore, trained on text-image pairs collected before 2021~\cite{hessel2021clipscore}, and Aesthetic Predictor, trained on the AVA dataset~\cite{murray2012ava}, tend to underperform when dealing with open-scene texts and images, often producing predictions inconsistent with human expectations~\cite{xu2023imagereward}. For example, the Aesthetic Predictor outputs values within the narrow range of $[4.8, 5.6]$, indicating its poor predictive ability within this data domain (the model's output range is $[0,10]$). X-IQE addresses this issue by incorporating domain-specific knowledge, such as scoring criteria, into the inference process. But unlike the unimodal nature of human evaluations, X-IQE's score distributions are bimodal, suggesting further room for improvement.

\subsection{Benchmarking SOTA Visual Generative Models}

\begin{table}[t]
\small
\centering
\caption{Benchmarking text-to-image models on COCO Caption.}
\setlength{\tabcolsep}{0.4mm} 
\renewcommand{\arraystretch}{1.3} 
\begin{tabular}{c|c|c|c|c|c|c|c|c}
\toprule
\multicolumn{1}{c|}{\multirow{2}{*}{Model}} &  \multicolumn{1}{c|}{\multirow{2}{*}{~~CLIP~~}} & \multicolumn{1}{c|}{\multirow{2}{*}{~Aes. Pred.~}} & \multicolumn{1}{c|}{\multirow{2}{*}{~ImgRwd~}} & \multicolumn{1}{c|}{\multirow{2}{*}{~~~~HPS~~~~}} & \multicolumn{4}{c}{X-IQE} \\
\multicolumn{1}{c|}{} & \multicolumn{1}{c|}{} & \multicolumn{1}{c|}{} & \multicolumn{1}{c|}{} & \multicolumn{1}{c|}{} & \multicolumn{1}{c|}{~Fidelity~~} & \multicolumn{1}{c|}{~Alignment~~} & \multicolumn{1}{c|}{~Aesthetics~~} & \multicolumn{1}{c}{~Overall~} \\
\hline
\hline
\rule{0pt}{4pt} 
~~Stable Diffusion 1.4~~ & 0.803 & 5.22 & 0.104 & 0.1966 & 5.47 & 3.29 & 5.76 & 14.52 \\
~~Stable Diffusion 2.1~~ & \textbf{0.831} & \textbf{5.42} & 0.472 & 0.1988 & 5.52 & 3.45 & 5.77 & 14.74 \\
~~Openjourney~~  & 0.806 & 5.38 & 0.244 & 0.1990 & 5.44 & 3.37 & \textbf{5.96} & 14.77 \\
~~DeepFloyd-IF~~  & 0.828 & 5.26 & \textbf{0.703} & \textbf{0.1994} & \textbf{5.55} & \textbf{3.52} & 5.79 & \textbf{14.86} \\
\bottomrule
\end{tabular}
\label{tab:COCO}
\vspace{-0.3cm} 
\end{table}

\begin{table}[t]
\small
\centering
\caption{Benchmarking text-to-image models on DrawBench.}
\setlength{\tabcolsep}{0.4mm} 
\renewcommand{\arraystretch}{1.3} 
\begin{tabular}{c|c|c|c|c|c|c|c|c}
\toprule
\multicolumn{1}{c|}{\multirow{2}{*}{Model}} &  \multicolumn{1}{c|}{\multirow{2}{*}{~~CLIP~~}} & \multicolumn{1}{c|}{\multirow{2}{*}{~Aes. Pred.~}} & \multicolumn{1}{c|}{\multirow{2}{*}{~ImgRwd~}} & \multicolumn{1}{c|}{\multirow{2}{*}{~~~~HPS~~~~}} & \multicolumn{4}{c}{X-IQE} \\
\multicolumn{1}{c|}{} & \multicolumn{1}{c|}{} & \multicolumn{1}{c|}{} & \multicolumn{1}{c|}{} & \multicolumn{1}{c|}{} & \multicolumn{1}{c|}{~Fidelity~~} & \multicolumn{1}{c|}{~Alignment~~} & \multicolumn{1}{c|}{~Aesthetics~~} & \multicolumn{1}{c}{~Overall~~} \\
\hline
\hline
\rule{0pt}{4pt} 
~~Stable Diffusion 1.4~~ & 0.793 & 5.09 & -0.029 & 0.1945 & \textbf{5.32} & 2.72 & 5.40 & 13.44 \\
~~Stable Diffusion 2.1~~ & 0.817 & 5.31 & 0.163 & 0.1955 & 5.10 & 2.50 & 5.04 & 12.64 \\
~~Openjourney~~  & 0.787 & \textbf{5.35} & 0.056 & 0.1972 & 5.14 & 2.62 & 5.21 & 12.97 \\
~~DeepFloyd-IF~~  & \textbf{0.827} & 5.10 & \textbf{0.541} & \textbf{0.1977} & \textbf{5.32} & \textbf{2.96} & \textbf{5.64} & \textbf{13.92} \\
\bottomrule
\end{tabular}
\label{tab:DrawBench}
\end{table}

One of the significant applications of X-IQE is its role as an impartial referee in evaluating various generative models. We evaluate four SOTA text-to-image models, including Stable Diffusion 1.4 and 2.1, Openjourney, and DeepFloyd-IF. Furthermore, we incorporate the results from two leading evaluation models specifically trained on AI-generated images (\ie ImageReward~\cite{xu2023imagereward} and HPS~\cite{wu2023better}). 

Tab. \ref{tab:COCO} displays the outcomes for images generated using the COCO Caption. Notably, CLIPScore and Aesthetic Predictor do not align human perception, whereas X-IQE exhibits strong agreement with ImageReward and HPS, selecting the best DeepFloyd-IF among the four methods. The overall score of X-IQE exhibits the same ranking as HPS. X-IQE also demonstrates its ability to extract reasonable detailed item scores. DeepFloyd-IF, serving as a public reproduction of Imagen~\cite{saharia2022photorealistic}, generates images in the pixel space guided by the text encoder T5 XXL~\cite{radford2019language}, enabling it to produce highly photorealistic images aligning well with the provided texts. In terms of aesthetics, Openjourney outperforms other models due to its high-quality training data generated by Midjourney.

\begin{figure}[t]
\vspace{0cm}                          
\centering\centerline{\includegraphics[width=1.0\linewidth]{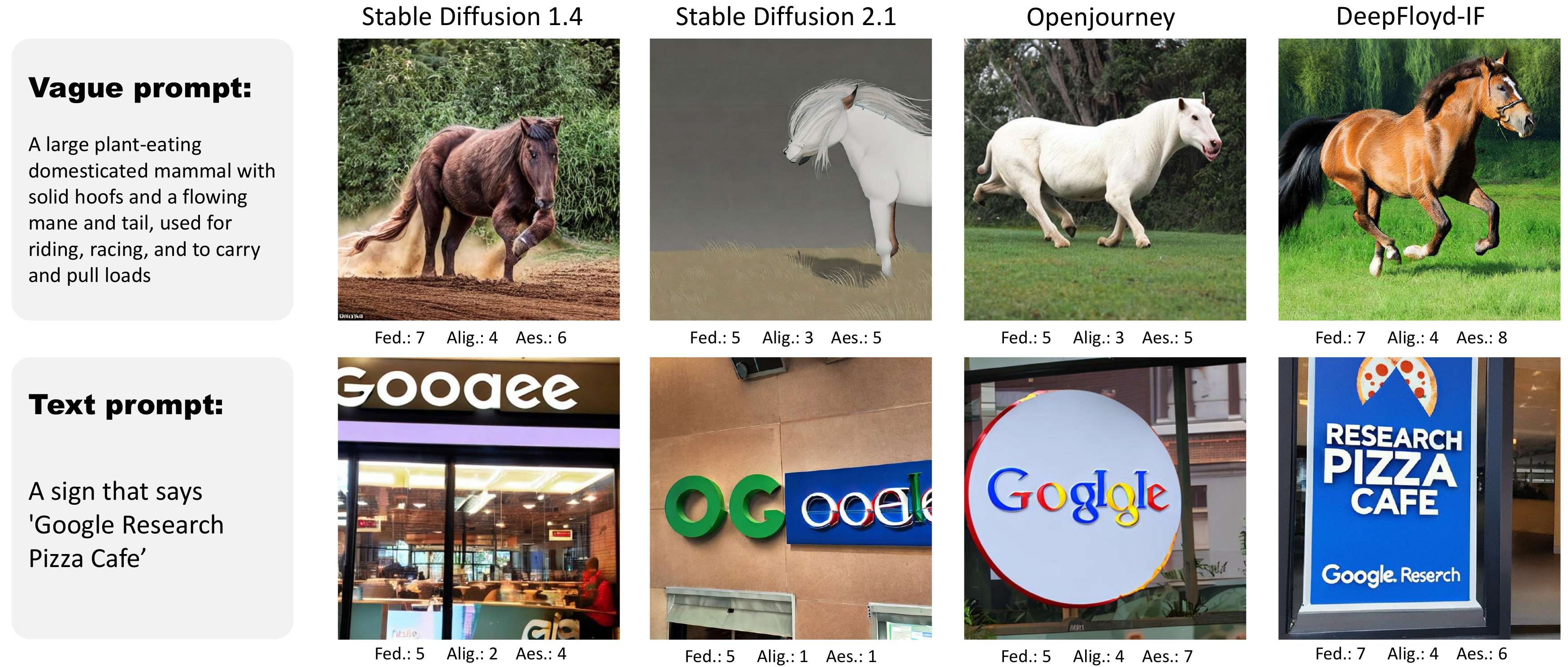}}
\caption{Comparison of images generated with DrawBench prompts and their X-IQE scoring.}
\label{fig:benchmark} 
\end{figure}

Tab. \ref{tab:DrawBench} shows the evaluation results on DrawBench, a challenging benchmark including unconventional prompts like vague descriptions and text generation. In this scenario, X-IQE provides conclusions that differ from those of ImageReward and HPS. Fig. \ref{fig:benchmark} illustrates the images generated with the four models. X-IQE accurately ranks them, especially for the poor alignment and aesthetics of SD2.1. This can be attributed to the robust capabilities of X-IQE in parsing long and ambiguous texts and recognizing texts within the images, which traditional models struggle to accomplish.

\section{Limitations}
\label{sect:limitations}

\textbf{Model capability.} Most visual encoders, including ViT in MiniGPT-4, were pre-trained using low-resolution images (\eg $224^2$), whereas the generated images using modern diffusion techniques often range from $512^2$ to $1024^2$. The significant downsampling of images during inference could negatively impact the model's ability to discriminate fine details. Additionally, it has been shown that CoTs typically work best for models with a size larger than 60B. When using smaller models like MiniGPT-4 with 7B/13B parameters, we observed various unexpected responses such as hallucinations, repeated answers, and incorrect output formats, especially with temperatures smaller than 0.1.

\textbf{CoT design.} The CoT strategy proposed in this work is concise and includes only necessary information for the LLM's inference. Longer and more detailed CoTs may be more effective but were not established due to the limited text length that current LLMs can support, typically shorter than 1,000 tokens. It has also been observed that X-IQE typically rates the image only with integers and tends to produce extreme quantitative results. This could be mitigated in future research by improving the prompts.

\section{Conclusion}

In this study, we have introduced X-IQE, the first explainable method for image quality evaluation with LLMs. This approach was developed to address the limitations of existing image evaluation models and to align the evaluation results more closely with human judgments. By employing a meticulously designed CoT strategy, X-IQE is able to differentiate AI-generated images, assess the alignment between text and image, and analyze aesthetic elements. Our experimental results demonstrate the reproducibility and effectiveness of X-IQE, as well as its capability to serve as a benchmark for current SOTA text-to-image generation methods without requiring additional training data. The ability of X-IQE to provide explanations for its reasoning process enables researchers to identify and target its weaknesses for further improvement efficiently.

\bibliographystyle{plainnat}

\bibliography{XIQE}

\begin{thebibliography}{56}
\providecommand{\natexlab}[1]{#1}
\providecommand{\url}[1]{\texttt{#1}}
\expandafter\ifx\csname urlstyle\endcsname\relax
  \providecommand{\doi}[1]{doi: #1}\else
  \providecommand{\doi}{doi: \begingroup \urlstyle{rm}\Url}\fi

\bibitem[Aguinis et~al.(2021)Aguinis, Villamor, and Ramani]{aguinis2021mturk}
Herman Aguinis, Isabel Villamor, and Ravi~S Ramani.
\newblock Mturk research: Review and recommendations.
\newblock \emph{Journal of Management}, 47\penalty0 (4):\penalty0 823--837,
  2021.

\bibitem[Bhattacharya et~al.(2011)Bhattacharya, Sukthankar, and
  Shah]{bhattacharya2011holistic}
Subhabrata Bhattacharya, Rahul Sukthankar, and Mubarak Shah.
\newblock A holistic approach to aesthetic enhancement of photographs.
\newblock \emph{ACM Transactions on Multimedia Computing, Communications, and
  Applications (TOMM)}, 7\penalty0 (1):\penalty0 1--21, 2011.

\bibitem[Brown et~al.(2020)Brown, Mann, Ryder, Subbiah, Kaplan, Dhariwal,
  Neelakantan, Shyam, Sastry, Askell, et~al.]{brown2020language}
Tom Brown, Benjamin Mann, Nick Ryder, Melanie Subbiah, Jared~D Kaplan, Prafulla
  Dhariwal, Arvind Neelakantan, Pranav Shyam, Girish Sastry, Amanda Askell,
  et~al.
\newblock Language models are few-shot learners.
\newblock \emph{Advances in neural information processing systems},
  33:\penalty0 1877--1901, 2020.

\bibitem[Bubeck et~al.(2023)Bubeck, Chandrasekaran, Eldan, Gehrke, Horvitz,
  Kamar, Lee, Lee, Li, Lundberg, et~al.]{bubeck2023sparks}
S{\'e}bastien Bubeck, Varun Chandrasekaran, Ronen Eldan, Johannes Gehrke, Eric
  Horvitz, Ece Kamar, Peter Lee, Yin~Tat Lee, Yuanzhi Li, Scott Lundberg,
  et~al.
\newblock Sparks of artificial general intelligence: Early experiments with
  gpt-4.
\newblock \emph{arXiv preprint arXiv:2303.12712}, 2023.

\bibitem[Chang et~al.(2017)Chang, Lu, and Chen]{chang2017aesthetic}
Kuang-Yu Chang, Kung-Hung Lu, and Chu-Song Chen.
\newblock Aesthetic critiques generation for photos.
\newblock In \emph{Proceedings of the IEEE international conference on computer
  vision}, pages 3514--3523, 2017.

\bibitem[Chen et~al.(2022)Chen, Guo, Yi, Li, and Elhoseiny]{chen2022visualgpt}
Jun Chen, Han Guo, Kai Yi, Boyang Li, and Mohamed Elhoseiny.
\newblock Visualgpt: Data-efficient adaptation of pretrained language models
  for image captioning.
\newblock In \emph{Proceedings of the IEEE/CVF Conference on Computer Vision
  and Pattern Recognition}, pages 18030--18040, 2022.

\bibitem[Chen et~al.(2015)Chen, Fang, Lin, Vedantam, Gupta, Doll{\'a}r, and
  Zitnick]{chen2015microsoft}
Xinlei Chen, Hao Fang, Tsung-Yi Lin, Ramakrishna Vedantam, Saurabh Gupta, Piotr
  Doll{\'a}r, and C~Lawrence Zitnick.
\newblock Microsoft coco captions: Data collection and evaluation server.
\newblock \emph{arXiv preprint arXiv:1504.00325}, 2015.

\bibitem[Chen et~al.(2017)Chen, Klopp, Sun, Chien, and Ma]{chen2017learning}
Yi-Ling Chen, Jan Klopp, Min Sun, Shao-Yi Chien, and Kwan-Liu Ma.
\newblock Learning to compose with professional photographs on the web.
\newblock In \emph{Proceedings of the 25th ACM international conference on
  Multimedia}, pages 37--45, 2017.

\bibitem[Chiang et~al.(2023)Chiang, Li, Lin, Sheng, Wu, Zhang, Zheng, Zhuang,
  Zhuang, Gonzalez, et~al.]{chiang2023vicuna}
Wei-Lin Chiang, Zhuohan Li, Zi~Lin, Ying Sheng, Zhanghao Wu, Hao Zhang, Lianmin
  Zheng, Siyuan Zhuang, Yonghao Zhuang, Joseph~E Gonzalez, et~al.
\newblock Vicuna: An open-source chatbot impressing gpt-4 with 90\%* chatgpt
  quality, 2023.

\bibitem[Chowdhery et~al.(2022)Chowdhery, Narang, Devlin, Bosma, Mishra,
  Roberts, Barham, Chung, Sutton, Gehrmann, et~al.]{chowdhery2022palm}
Aakanksha Chowdhery, Sharan Narang, Jacob Devlin, Maarten Bosma, Gaurav Mishra,
  Adam Roberts, Paul Barham, Hyung~Won Chung, Charles Sutton, Sebastian
  Gehrmann, et~al.
\newblock Palm: Scaling language modeling with pathways.
\newblock \emph{arXiv preprint arXiv:2204.02311}, 2022.

\bibitem[Chung et~al.(2022)Chung, Hou, Longpre, Zoph, Tay, Fedus, Li, Wang,
  Dehghani, Brahma, et~al.]{chung2022scaling}
Hyung~Won Chung, Le~Hou, Shayne Longpre, Barret Zoph, Yi~Tay, William Fedus,
  Eric Li, Xuezhi Wang, Mostafa Dehghani, Siddhartha Brahma, et~al.
\newblock Scaling instruction-finetuned language models.
\newblock \emph{arXiv preprint arXiv:2210.11416}, 2022.

\bibitem[Devlin et~al.(2018)Devlin, Chang, Lee, and Toutanova]{devlin2018bert}
Jacob Devlin, Ming-Wei Chang, Kenton Lee, and Kristina Toutanova.
\newblock Bert: Pre-training of deep bidirectional transformers for language
  understanding.
\newblock \emph{arXiv preprint arXiv:1810.04805}, 2018.

\bibitem[Ding et~al.(2022)Ding, Zheng, Hong, and Tang]{ding2022cogview2}
Ming Ding, Wendi Zheng, Wenyi Hong, and Jie Tang.
\newblock Cogview2: Faster and better text-to-image generation via hierarchical
  transformers.
\newblock \emph{arXiv preprint arXiv:2204.14217}, 2022.

\bibitem[Dosovitskiy et~al.(2020)Dosovitskiy, Beyer, Kolesnikov, Weissenborn,
  Zhai, Unterthiner, Dehghani, Minderer, Heigold, Gelly,
  et~al.]{dosovitskiy2020image}
Alexey Dosovitskiy, Lucas Beyer, Alexander Kolesnikov, Dirk Weissenborn,
  Xiaohua Zhai, Thomas Unterthiner, Mostafa Dehghani, Matthias Minderer, Georg
  Heigold, Sylvain Gelly, et~al.
\newblock An image is worth 16x16 words: Transformers for image recognition at
  scale.
\newblock \emph{arXiv preprint arXiv:2010.11929}, 2020.

\bibitem[Goodfellow et~al.(2020)Goodfellow, Pouget-Abadie, Mirza, Xu,
  Warde-Farley, Ozair, Courville, and Bengio]{goodfellow2020generative}
Ian Goodfellow, Jean Pouget-Abadie, Mehdi Mirza, Bing Xu, David Warde-Farley,
  Sherjil Ozair, Aaron Courville, and Yoshua Bengio.
\newblock Generative adversarial networks.
\newblock \emph{Communications of the ACM}, 63\penalty0 (11):\penalty0
  139--144, 2020.

\bibitem[Hessel et~al.(2021)Hessel, Holtzman, Forbes, Le~Bras, and
  Choi]{hessel2021clipscore}
Jack Hessel, Ari Holtzman, Maxwell Forbes, Ronan Le~Bras, and Yejin Choi.
\newblock Clipscore: A reference-free evaluation metric for image captioning.
\newblock In \emph{Proceedings of the 2021 Conference on Empirical Methods in
  Natural Language Processing}, pages 7514--7528, 2021.

\bibitem[Heusel et~al.(2017)Heusel, Ramsauer, Unterthiner, Nessler, and
  Hochreiter]{heusel2017gans}
Martin Heusel, Hubert Ramsauer, Thomas Unterthiner, Bernhard Nessler, and Sepp
  Hochreiter.
\newblock Gans trained by a two time-scale update rule converge to a local nash
  equilibrium.
\newblock \emph{Advances in neural information processing systems}, 30, 2017.

\bibitem[Ho et~al.(2020)Ho, Jain, and Abbeel]{ho2020denoising}
Jonathan Ho, Ajay Jain, and Pieter Abbeel.
\newblock Denoising diffusion probabilistic models.
\newblock \emph{Advances in Neural Information Processing Systems},
  33:\penalty0 6840--6851, 2020.

\bibitem[Hoffmann et~al.(2022)Hoffmann, Borgeaud, Mensch, Buchatskaya, Cai,
  Rutherford, Casas, Hendricks, Welbl, Clark, et~al.]{hoffmann2022training}
Jordan Hoffmann, Sebastian Borgeaud, Arthur Mensch, Elena Buchatskaya, Trevor
  Cai, Eliza Rutherford, Diego de~Las Casas, Lisa~Anne Hendricks, Johannes
  Welbl, Aidan Clark, et~al.
\newblock Training compute-optimal large language models.
\newblock \emph{arXiv preprint arXiv:2203.15556}, 2022.

\bibitem[Hu et~al.(2023)Hu, Chen, Xiao, Sun, Chen, Yuille, and
  Zhou]{hu2023label}
Qixin Hu, Yixiong Chen, Junfei Xiao, Shuwen Sun, Jieneng Chen, Alan Yuille, and
  Zongwei Zhou.
\newblock Label-free liver tumor segmentation.
\newblock \emph{arXiv preprint arXiv:2303.14869}, 2023.

\bibitem[Kennedy et~al.(2020)Kennedy, Clifford, Burleigh, Waggoner, Jewell, and
  Winter]{kennedy2020shape}
Ryan Kennedy, Scott Clifford, Tyler Burleigh, Philip~D Waggoner, Ryan Jewell,
  and Nicholas~JG Winter.
\newblock The shape of and solutions to the mturk quality crisis.
\newblock \emph{Political Science Research and Methods}, 8\penalty0
  (4):\penalty0 614--629, 2020.

\bibitem[Khosla et~al.(2015)Khosla, Raju, Torralba, and
  Oliva]{khosla2015understanding}
Aditya Khosla, Akhil~S Raju, Antonio Torralba, and Aude Oliva.
\newblock Understanding and predicting image memorability at a large scale.
\newblock In \emph{Proceedings of the IEEE international conference on computer
  vision}, pages 2390--2398, 2015.

\bibitem[Krippendorff(2018)]{krippendorff2018content}
Klaus Krippendorff.
\newblock \emph{Content analysis: An introduction to its methodology}.
\newblock Sage publications, 2018.

\bibitem[Kynk{\"a}{\"a}nniemi et~al.(2019)Kynk{\"a}{\"a}nniemi, Karras, Laine,
  Lehtinen, and Aila]{kynkaanniemi2019improved}
Tuomas Kynk{\"a}{\"a}nniemi, Tero Karras, Samuli Laine, Jaakko Lehtinen, and
  Timo Aila.
\newblock Improved precision and recall metric for assessing generative models.
\newblock \emph{Advances in Neural Information Processing Systems}, 32, 2019.

\bibitem[Lee et~al.(2023)Lee, Liu, Ryu, Watkins, Du, Boutilier, Abbeel,
  Ghavamzadeh, and Gu]{lee2023aligning}
Kimin Lee, Hao Liu, Moonkyung Ryu, Olivia Watkins, Yuqing Du, Craig Boutilier,
  Pieter Abbeel, Mohammad Ghavamzadeh, and Shixiang~Shane Gu.
\newblock Aligning text-to-image models using human feedback.
\newblock \emph{arXiv preprint arXiv:2302.12192}, 2023.

\bibitem[Li et~al.(2022)Li, Li, Xiong, and Hoi]{li2022blip}
Junnan Li, Dongxu Li, Caiming Xiong, and Steven Hoi.
\newblock Blip: Bootstrapping language-image pre-training for unified
  vision-language understanding and generation.
\newblock In \emph{International Conference on Machine Learning}, pages
  12888--12900. PMLR, 2022.

\bibitem[Li et~al.(2023)Li, Li, Savarese, and Hoi]{li2023blip}
Junnan Li, Dongxu Li, Silvio Savarese, and Steven Hoi.
\newblock Blip-2: Bootstrapping language-image pre-training with frozen image
  encoders and large language models.
\newblock \emph{arXiv preprint arXiv:2301.12597}, 2023.

\bibitem[Liu et~al.(2020)Liu, Puri, Kamath, and
  Bhattacharya]{liu2020composition}
Dong Liu, Rohit Puri, Nagendra Kamath, and Subhabrata Bhattacharya.
\newblock Composition-aware image aesthetics assessment.
\newblock In \emph{Proceedings of the IEEE/CVF Winter Conference on
  Applications of Computer Vision}, pages 3569--3578, 2020.

\bibitem[Lu et~al.(2022)Lu, Zhou, Bao, Chen, Li, and Zhu]{ludpm}
Cheng Lu, Yuhao Zhou, Fan Bao, Jianfei Chen, Chongxuan Li, and Jun Zhu.
\newblock Dpm-solver: A fast ode solver for diffusion probabilistic model
  sampling in around 10 steps.
\newblock In \emph{Advances in Neural Information Processing Systems}, 2022.

\bibitem[Lu et~al.(2015)Lu, Lin, Shen, Mech, and Wang]{lu2015deep}
Xin Lu, Zhe Lin, Xiaohui Shen, Radomir Mech, and James~Z Wang.
\newblock Deep multi-patch aggregation network for image style, aesthetics, and
  quality estimation.
\newblock In \emph{Proceedings of the IEEE international conference on computer
  vision}, pages 990--998, 2015.

\bibitem[Min et~al.(2022)Min, Lyu, Holtzman, Artetxe, Lewis, Hajishirzi, and
  Zettlemoyer]{min2022rethinking}
Sewon Min, Xinxi Lyu, Ari Holtzman, Mikel Artetxe, Mike Lewis, Hannaneh
  Hajishirzi, and Luke Zettlemoyer.
\newblock Rethinking the role of demonstrations: What makes in-context learning
  work?
\newblock \emph{arXiv preprint arXiv:2202.12837}, 2022.

\bibitem[Murray et~al.(2012)Murray, Marchesotti, and Perronnin]{murray2012ava}
Naila Murray, Luca Marchesotti, and Florent Perronnin.
\newblock Ava: A large-scale database for aesthetic visual analysis.
\newblock In \emph{2012 IEEE conference on computer vision and pattern
  recognition}, pages 2408--2415. IEEE, 2012.

\bibitem[Nichol and Dhariwal(2021)]{nichol2021improved}
Alexander~Quinn Nichol and Prafulla Dhariwal.
\newblock Improved denoising diffusion probabilistic models.
\newblock In \emph{International Conference on Machine Learning}, pages
  8162--8171. PMLR, 2021.

\bibitem[Nichol et~al.(2022)Nichol, Dhariwal, Ramesh, Shyam, Mishkin, Mcgrew,
  Sutskever, and Chen]{nichol2022glide}
Alexander~Quinn Nichol, Prafulla Dhariwal, Aditya Ramesh, Pranav Shyam, Pamela
  Mishkin, Bob Mcgrew, Ilya Sutskever, and Mark Chen.
\newblock Glide: Towards photorealistic image generation and editing with
  text-guided diffusion models.
\newblock In \emph{International Conference on Machine Learning}, pages
  16784--16804. PMLR, 2022.

\bibitem[Otani et~al.(2023)Otani, Togashi, Sawai, Ishigami, Nakashima, Rahtu,
  Heikkil{\"a}, and Satoh]{otani2023toward}
Mayu Otani, Riku Togashi, Yu~Sawai, Ryosuke Ishigami, Yuta Nakashima, Esa
  Rahtu, Janne Heikkil{\"a}, and Shin'ichi Satoh.
\newblock Toward verifiable and reproducible human evaluation for text-to-image
  generation.
\newblock \emph{arXiv preprint arXiv:2304.01816}, 2023.

\bibitem[Ouyang et~al.(2022)Ouyang, Wu, Jiang, Almeida, Wainwright, Mishkin,
  Zhang, Agarwal, Slama, Ray, et~al.]{ouyang2022training}
Long Ouyang, Jeffrey Wu, Xu~Jiang, Diogo Almeida, Carroll Wainwright, Pamela
  Mishkin, Chong Zhang, Sandhini Agarwal, Katarina Slama, Alex Ray, et~al.
\newblock Training language models to follow instructions with human feedback.
\newblock \emph{Advances in Neural Information Processing Systems},
  35:\penalty0 27730--27744, 2022.

\bibitem[Palmer et~al.(2013)Palmer, Schloss, and Sammartino]{palmer2013visual}
Stephen~E Palmer, Karen~B Schloss, and Jonathan Sammartino.
\newblock Visual aesthetics and human preference.
\newblock \emph{Annual review of psychology}, 64:\penalty0 77--107, 2013.

\bibitem[Radford et~al.(2019)Radford, Wu, Child, Luan, Amodei, Sutskever,
  et~al.]{radford2019language}
Alec Radford, Jeffrey Wu, Rewon Child, David Luan, Dario Amodei, Ilya
  Sutskever, et~al.
\newblock Language models are unsupervised multitask learners.
\newblock \emph{OpenAI blog}, 1\penalty0 (8):\penalty0 9, 2019.

\bibitem[Radford et~al.(2021)Radford, Kim, Hallacy, Ramesh, Goh, Agarwal,
  Sastry, Askell, Mishkin, Clark, et~al.]{radford2021learning}
Alec Radford, Jong~Wook Kim, Chris Hallacy, Aditya Ramesh, Gabriel Goh,
  Sandhini Agarwal, Girish Sastry, Amanda Askell, Pamela Mishkin, Jack Clark,
  et~al.
\newblock Learning transferable visual models from natural language
  supervision.
\newblock In \emph{International conference on machine learning}, pages
  8748--8763. PMLR, 2021.

\bibitem[Raffel et~al.(2020)Raffel, Shazeer, Roberts, Lee, Narang, Matena,
  Zhou, Li, and Liu]{raffel2020exploring}
Colin Raffel, Noam Shazeer, Adam Roberts, Katherine Lee, Sharan Narang, Michael
  Matena, Yanqi Zhou, Wei Li, and Peter~J Liu.
\newblock Exploring the limits of transfer learning with a unified text-to-text
  transformer.
\newblock \emph{The Journal of Machine Learning Research}, 21\penalty0
  (1):\penalty0 5485--5551, 2020.

\bibitem[Rombach et~al.(2022)Rombach, Blattmann, Lorenz, Esser, and
  Ommer]{rombach2022high}
Robin Rombach, Andreas Blattmann, Dominik Lorenz, Patrick Esser, and Bj{\"o}rn
  Ommer.
\newblock High-resolution image synthesis with latent diffusion models.
\newblock In \emph{Proceedings of the IEEE/CVF Conference on Computer Vision
  and Pattern Recognition}, pages 10684--10695, 2022.

\bibitem[Saharia et~al.(2022)Saharia, Chan, Saxena, Li, Whang, Denton,
  Ghasemipour, Gontijo~Lopes, Karagol~Ayan, Salimans,
  et~al.]{saharia2022photorealistic}
Chitwan Saharia, William Chan, Saurabh Saxena, Lala Li, Jay Whang, Emily~L
  Denton, Kamyar Ghasemipour, Raphael Gontijo~Lopes, Burcu Karagol~Ayan, Tim
  Salimans, et~al.
\newblock Photorealistic text-to-image diffusion models with deep language
  understanding.
\newblock \emph{Advances in Neural Information Processing Systems},
  35:\penalty0 36479--36494, 2022.

\bibitem[Salimans et~al.(2016)Salimans, Goodfellow, Zaremba, Cheung, Radford,
  and Chen]{salimans2016improved}
Tim Salimans, Ian Goodfellow, Wojciech Zaremba, Vicki Cheung, Alec Radford, and
  Xi~Chen.
\newblock Improved techniques for training gans.
\newblock \emph{Advances in neural information processing systems}, 29, 2016.

\bibitem[Schuhmann et~al.(2022)Schuhmann, Beaumont, Vencu, Gordon, Wightman,
  Cherti, Coombes, Katta, Mullis, Wortsman, et~al.]{schuhmannlaion}
Christoph Schuhmann, Romain Beaumont, Richard Vencu, Cade~W Gordon, Ross
  Wightman, Mehdi Cherti, Theo Coombes, Aarush Katta, Clayton Mullis, Mitchell
  Wortsman, et~al.
\newblock Laion-5b: An open large-scale dataset for training next generation
  image-text models.
\newblock In \emph{Thirty-sixth Conference on Neural Information Processing
  Systems Datasets and Benchmarks Track}, 2022.

\bibitem[Sheng et~al.(2018)Sheng, Dong, Ma, Mei, Huang, and
  Hu]{sheng2018attention}
Kekai Sheng, Weiming Dong, Chongyang Ma, Xing Mei, Feiyue Huang, and Bao-Gang
  Hu.
\newblock Attention-based multi-patch aggregation for image aesthetic
  assessment.
\newblock In \emph{Proceedings of the 26th ACM international conference on
  Multimedia}, pages 879--886, 2018.

\bibitem[Smith et~al.(2022)Smith, Patwary, Norick, LeGresley, Rajbhandari,
  Casper, Liu, Prabhumoye, Zerveas, Korthikanti, et~al.]{smith2022using}
Shaden Smith, Mostofa Patwary, Brandon Norick, Patrick LeGresley, Samyam
  Rajbhandari, Jared Casper, Zhun Liu, Shrimai Prabhumoye, George Zerveas,
  Vijay Korthikanti, et~al.
\newblock Using deepspeed and megatron to train megatron-turing nlg 530b, a
  large-scale generative language model.
\newblock \emph{arXiv preprint arXiv:2201.11990}, 2022.

\bibitem[Song et~al.(2020)Song, Sohl-Dickstein, Kingma, Kumar, Ermon, and
  Poole]{song2020score}
Yang Song, Jascha Sohl-Dickstein, Diederik~P Kingma, Abhishek Kumar, Stefano
  Ermon, and Ben Poole.
\newblock Score-based generative modeling through stochastic differential
  equations.
\newblock \emph{arXiv preprint arXiv:2011.13456}, 2020.

\bibitem[Touvron et~al.(2023)Touvron, Lavril, Izacard, Martinet, Lachaux,
  Lacroix, Rozi{\`e}re, Goyal, Hambro, Azhar, et~al.]{touvron2023llama}
Hugo Touvron, Thibaut Lavril, Gautier Izacard, Xavier Martinet, Marie-Anne
  Lachaux, Timoth{\'e}e Lacroix, Baptiste Rozi{\`e}re, Naman Goyal, Eric
  Hambro, Faisal Azhar, et~al.
\newblock Llama: Open and efficient foundation language models.
\newblock \emph{arXiv preprint arXiv:2302.13971}, 2023.

\bibitem[Tsimpoukelli et~al.(2021)Tsimpoukelli, Menick, Cabi, Eslami, Vinyals,
  and Hill]{tsimpoukelli2021multimodal}
Maria Tsimpoukelli, Jacob~L Menick, Serkan Cabi, SM~Eslami, Oriol Vinyals, and
  Felix Hill.
\newblock Multimodal few-shot learning with frozen language models.
\newblock \emph{Advances in Neural Information Processing Systems},
  34:\penalty0 200--212, 2021.

\bibitem[Wei et~al.(2022)Wei, Wang, Schuurmans, Bosma, Xia, Chi, Le, Zhou,
  et~al.]{weichain}
Jason Wei, Xuezhi Wang, Dale Schuurmans, Maarten Bosma, Fei Xia, Ed~H Chi,
  Quoc~V Le, Denny Zhou, et~al.
\newblock Chain-of-thought prompting elicits reasoning in large language
  models.
\newblock In \emph{Advances in Neural Information Processing Systems}, 2022.

\bibitem[Wu et~al.(2023)Wu, Sun, Zhu, Zhao, and Li]{wu2023better}
Xiaoshi Wu, Keqiang Sun, Feng Zhu, Rui Zhao, and Hongsheng Li.
\newblock Better aligning text-to-image models with human preference.
\newblock \emph{arXiv preprint arXiv:2303.14420}, 2023.

\bibitem[Xu et~al.(2023)Xu, Liu, Wu, Tong, Li, Ding, Tang, and
  Dong]{xu2023imagereward}
Jiazheng Xu, Xiao Liu, Yuchen Wu, Yuxuan Tong, Qinkai Li, Ming Ding, Jie Tang,
  and Yuxiao Dong.
\newblock Imagereward: Learning and evaluating human preferences for
  text-to-image generation.
\newblock \emph{arXiv preprint arXiv:2304.05977}, 2023.

\bibitem[Yu et~al.(2022)Yu, Xu, Koh, Luong, Baid, Wang, Vasudevan, Ku, Yang,
  Ayan, et~al.]{yu2022scaling}
Jiahui Yu, Yuanzhong Xu, Jing~Yu Koh, Thang Luong, Gunjan Baid, Zirui Wang,
  Vijay Vasudevan, Alexander Ku, Yinfei Yang, Burcu~Karagol Ayan, et~al.
\newblock Scaling autoregressive models for content-rich text-to-image
  generation.
\newblock \emph{arXiv preprint arXiv:2206.10789}, 2022.

\bibitem[Zhang et~al.(2021)Zhang, Niu, and Zhang]{zhang2021image}
Bo~Zhang, Li~Niu, and Liqing Zhang.
\newblock Image composition assessment with saliency-augmented multi-pattern
  pooling.
\newblock \emph{arXiv preprint arXiv:2104.03133}, 2021.

\bibitem[Zhao et~al.(2023)Zhao, Zhou, Li, Tang, Wang, Hou, Min, Zhang, Zhang,
  Dong, et~al.]{zhao2023survey}
Wayne~Xin Zhao, Kun Zhou, Junyi Li, Tianyi Tang, Xiaolei Wang, Yupeng Hou,
  Yingqian Min, Beichen Zhang, Junjie Zhang, Zican Dong, et~al.
\newblock A survey of large language models.
\newblock \emph{arXiv preprint arXiv:2303.18223}, 2023.

\bibitem[Zhu et~al.(2023)Zhu, Chen, Shen, Li, and Elhoseiny]{zhu2023minigpt}
Deyao Zhu, Jun Chen, Xiaoqian Shen, Xiang Li, and Mohamed Elhoseiny.
\newblock Minigpt-4: Enhancing vision-language understanding with advanced
  large language models.
\newblock \emph{arXiv preprint arXiv:2304.10592}, 2023.

\end{thebibliography}

\newpage
\appendix
\section{All Prompts Used in the Experiments}

\begin{table}[!htbp]
\small
\centering
\caption{The main prompts used for X-IQE.}
\setlength{\tabcolsep}{1pt}
\renewcommand{\arraystretch}{1.3} 
\begin{tabularx}{\textwidth}{lX}
\toprule
\makebox[0.2\textwidth][c]{Description} & \makebox[0.8\textwidth][c]{Prompt} \\
\hline
\hline
\rule{0pt}{4pt} 
Fidelity Question & You are my assistant to evaluate the image quality. Briefly describe (within 50 words) the type (e.g., photo, painting) and content of this image, and analyze whether this image meets the following conditions of an AI-generated image (within 30 words per point).\newline\newline
1. Imperfect details: distorted, blurry, or irrational faces, limbs, fingers, objects, or texts.\newline
2. Improper composition: some misplaced object relationships.\newline
3. Strange colors: overly bright, saturated colors.\newline
4. Artificial look: looks like a real image but has an unclear rendering or other artificial look.\newline\newline
Provide your analysis in JSON format with the following keys: Image description, Imperfect details, Improper composition, Strange colors, Artificial look, Fidelity (e.g., 6/10). The fidelity scoring criteria are as follows:\newline\newline
Definitely AI-generated (0-1)\newline
Very likely AI-generated (2-3)\newline
Certain probability of AI generation (4)\newline
Unsure (5)\newline
Certain probability being real (6)\newline
Very real (7-8)\newline
Definitely real (9-10)\\
\hline
Alignment Question & According to the image and your previous description, how well does the image align with the following description?
\newline\newline<Image Caption>\newline\newline 
not match at all (1)\newline
Has significant discrepancies (2)\newline
Has several minor discrepancies (3)\newline
Has a few minor discrepancies (4)\newline
Matches exactly (5)\newline\newline
Provide your analysis in JSON format with the following keys: Alignment analysis (within 100 words), Alignment score (e.g., 4/5). 
\\
\hline
Aesthetics Question & Briefly analyze the aesthetic elements of this image (each item within 20 words) and score its aesthetics. The scoring criteria for each item are as follows.\newline\newline
Extremely bad (0-1)\newline
Poor quality (2-3)\newline
Below average (4)\newline
Average (5)\newline
Above average (6)\newline
Good (7-8)\newline
Excellent (9)\newline
Wonderful (10)\newline\newline
Provide your analysis in JSON format with the following keys: Color harmony, Color brightness, Color saturation, Composition, Perspective, Light and shadow, Detailed expression, Vivid posture, Visual impact, Overall aesthetic score (e.g., 6/10). \\
\bottomrule
\label{tab:prompts}
\end{tabularx}
\end{table}

\begin{table}[!htbp]
\small
\centering
\caption{The ablation prompts used for X-IQE.}
\setlength{\tabcolsep}{1pt}
\renewcommand{\arraystretch}{1.3} 
\begin{tabularx}{\textwidth}{lX}
\toprule
\makebox[0.2\textwidth][c]{Description} & \makebox[0.8\textwidth][c]{Prompt} \\
\hline
\hline
\rule{0pt}{4pt} 
Fidelity question baseline & You are my assistant to evaluate the image quality. Whether this image seems like a real image instead of AI-generated? Give your fidelity score out of 10 in a format like n/10.\\
\hline
Fidelity question w/o format & You are my assistant to evaluate the image quality. Whether this image seems like a real image instead of AI-generated? Give your fidelity score out of 10 in a format like n/10. Consider the following factor for AI images.\newline\newline
1. Imperfect details: distorted, blurry, or irrational faces, limbs, fingers, objects, or texts.\newline
2. Improper composition: some misplaced object relationships.\newline\
3. Strange colors: overly bright, saturated colors.\newline
4. Artificial look: looks like a real image but has an unclear rendering or other artificial look.\newline\newline
The fidelity scoring criteria are as follows:\newline\newline
Definitely AI-generated (0-1)\newline
Very likely AI-generated (2-3)\newline
Certain probability of AI generation (4)\newline
Unsure (5)\newline
Certain probability being real (6)\newline
Very real (7-8)\newline
Definitely real (9-10)\\
\hline
Alignment question baseline & How well does the image align with the following description?\newline\newline
<Image Caption>\newline\newline
Give an alignment score out of 5 like 3/5.\\
\hline
Alignment question w/o format & Briefly describe (within 50 words) the type (e.g., photo, painting) and content of this image, how well does the image align with the following description?\newline\newline
<Image Caption>\newline\newline
Does not match at all (1).\newline
Has significant discrepancies (2)\newline
Has several minor discrepancies (3)\newline
Has a few minor discrepancies (4)\newline
Matches exactly (5)\newline\newline
Give an alignment score out of 5 like 3/5.\\
\hline
Aesthetic question baseline & Briefly analyze the aesthetic elements of this image. Give an aesthetic score out of 10 like 6/10.\\
\hline
Aesthetic question w/o format & Briefly analyze the aesthetic elements (color harmony, color brightness, color saturation, composition, perspective, light and shadow, detailed expression, vivid posture, visual impact) of this image (each item within 20 words). Give an aesthetic score out of 10 like 6/10. The scoring criteria for each item are as follows.\newline\newline
Extremely bad (0-1)\newline
Poor quality (2-3)\newline
Below average (4)\newline
Average (5)\newline
Above average (6)\newline
Good (7-8)\newline
Excellent (9)\newline
Wonderful (10)\\
\bottomrule
\label{tab:ablation_prompts}
\end{tabularx}
\end{table}

\begin{table}[!tbp]
\small
\centering
\caption{The continue prompts used for X-IQE.}
\setlength{\tabcolsep}{1pt}
\renewcommand{\arraystretch}{1.3} 
\begin{tabularx}{\textwidth}{lX}
\toprule
\makebox[0.2\textwidth][c]{Description} & \makebox[0.8\textwidth][c]{Prompt} \\
\hline
\hline
\rule{0pt}{4pt} 
Fidelity Question (cont.) & Give the fidelity rating (in a format like n/10), do not repeat what you have said.\\
\hline
Alignment Question (cont.) & Give the rating of text-image alignment (in a format like n/5), do not repeat what you have said. \\
\hline
Aesthetics Question (cont.) & Give the rating of aesthetics (in a format like n/10), do not repeat what you have said.\\
\bottomrule
\label{tab:cont_prompts}
\end{tabularx}
\end{table}

\newpage\newpage\newpage
\section{Failure Situations}

\begin{figure}[ht]
\vspace{0cm}                          
\centering\centerline{\includegraphics[width=1.0\linewidth]{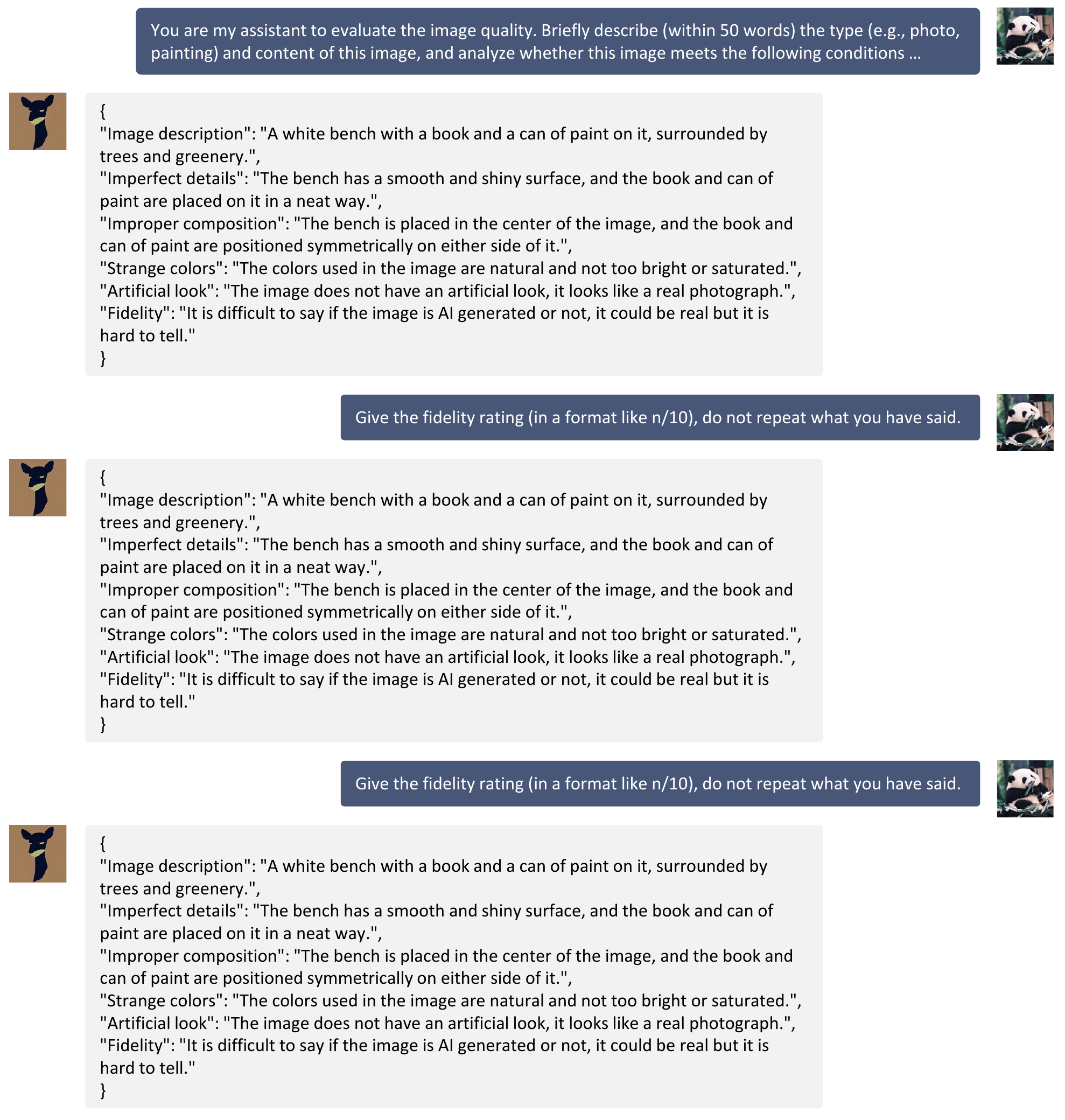}}
\caption{The example of repeating answers without scoring, which is a common bug of Vicuna.}
\label{fig:error2} 
\end{figure}

\begin{figure}[ht]
\vspace{0cm}                          
\centering\centerline{\includegraphics[width=1.0\linewidth]{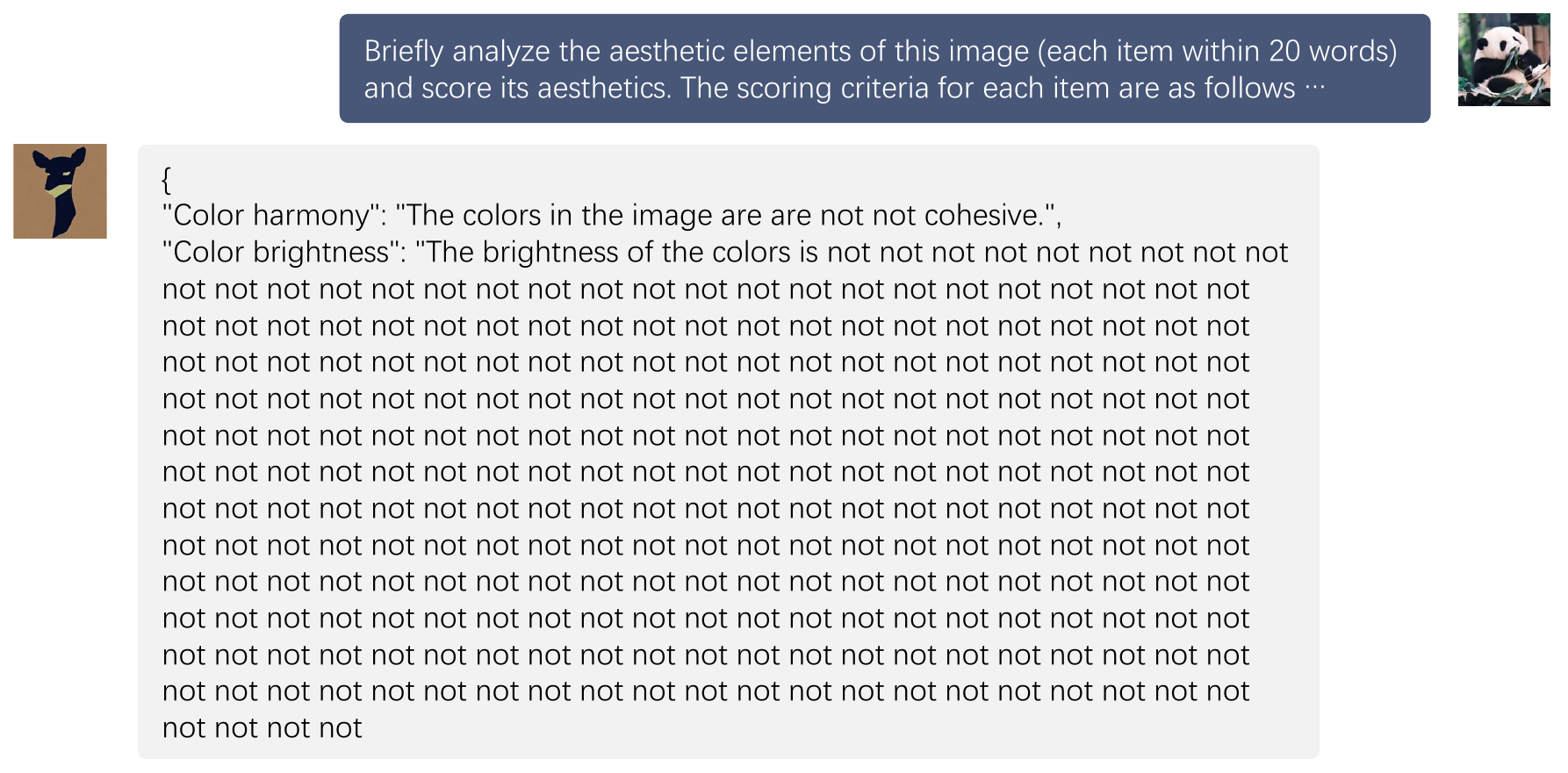}}
\caption{The example of repeating a single word. This situation often happens when the temperature is lower than 0.1.}
\label{fig:error3} 
\end{figure}

\begin{figure}[ht]
\vspace{0cm}                          
\centering\centerline{\includegraphics[width=1.0\linewidth]{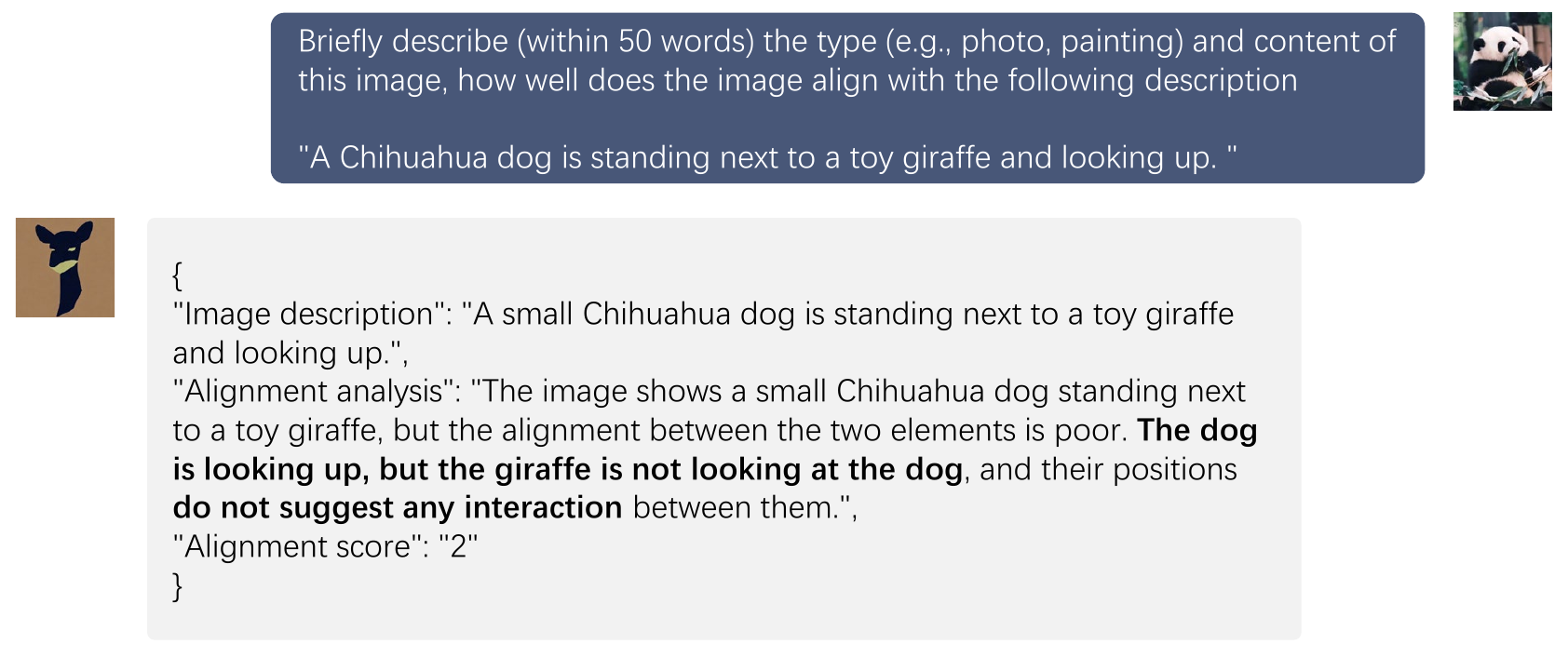}}
\caption{The example of hallucination error. The image caption does not mention the giraffe is looking at the dog or any of their interaction.}
\label{fig:error1} 
\end{figure}

\begin{figure}[ht]
\vspace{0cm}                          
\centering\centerline{\includegraphics[width=1.0\linewidth]{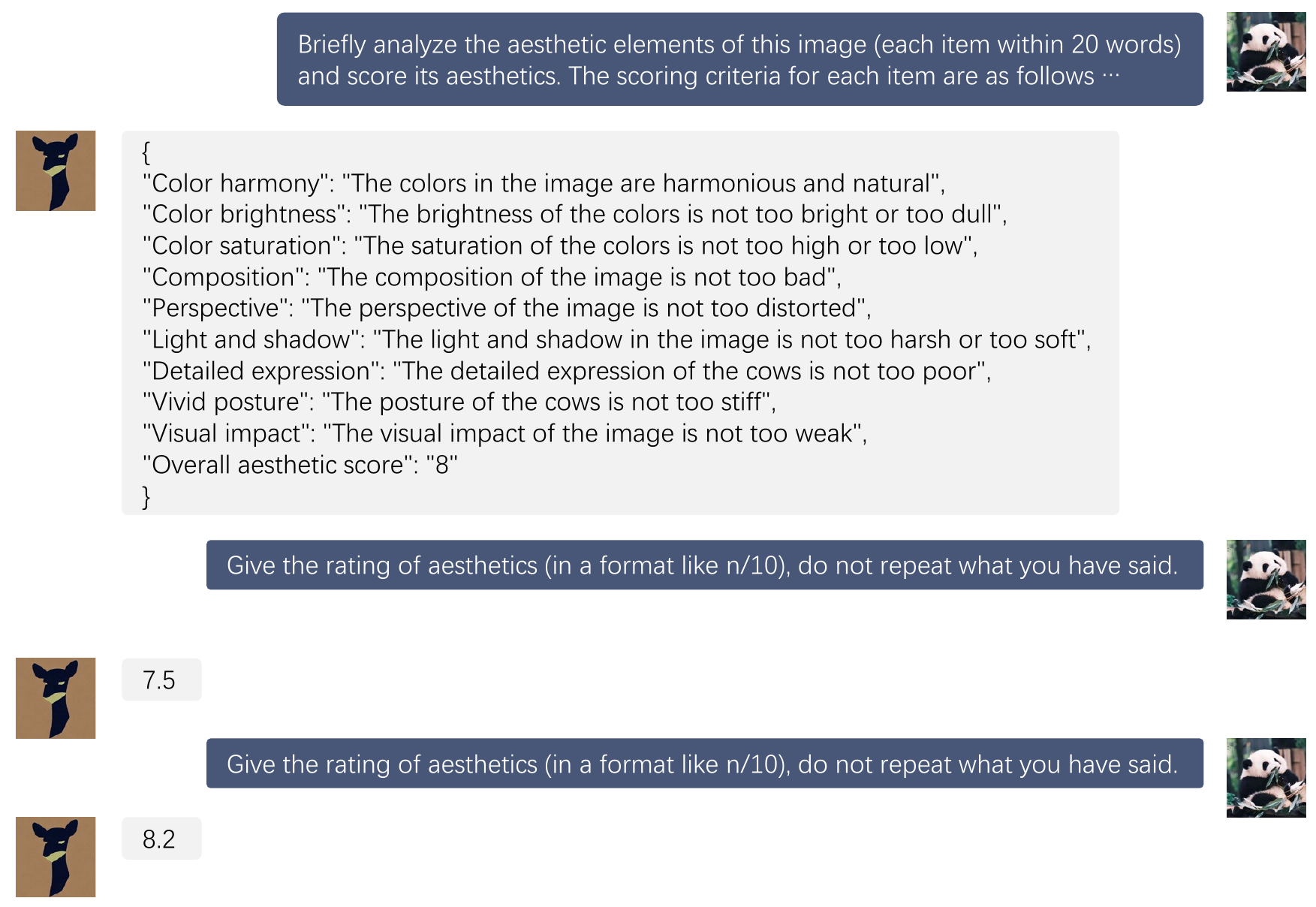}}
\caption{The example of inconsistent multiple responses.}
\label{fig:error4} 
\end{figure}

\begin{figure}[ht]
\vspace{0cm}                          
\centering\centerline{\includegraphics[width=1.0\linewidth]{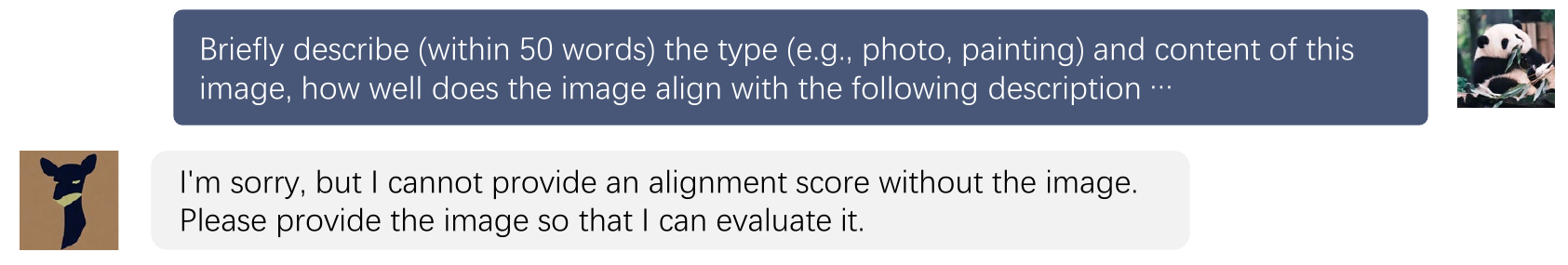}}
\caption{The example of giving no answer. This situation sometimes happens for the separate alignment evaluation. If fidelity evaluation had been conducted previously, the error would not happen.}
\label{fig:error5} 
\end{figure}


\end{document}